%% file: main.tex
\documentclass[10pt,twocolumn,letterpaper,dvipsnames]{article}

\usepackage[accsupp]{axessibility}  % Improves PDF readability for those with disabilities.
\usepackage{wacv}
\usepackage{times}
\usepackage{epsfig}
\usepackage{graphicx}
\usepackage{amsmath}
\usepackage{amssymb}
\usepackage{colortbl}
\usepackage{float}
\usepackage{multirow}
\usepackage{multicol}
\usepackage[table]{xcolor}
\usepackage{makecell}
\usepackage{booktabs}
\usepackage{subcaption}
\usepackage{hhline}
\usepackage{xcolor}
\usepackage[export]{adjustbox}
\usepackage{placeins}
\usepackage{authblk}

\usepackage{lipsum}

\definecolor{airforceblue}{rgb}{0.36, 0.54, 0.66}
\definecolor{cobalt}{rgb}{0.0, 0.28, 0.67}

\newcommand{\comm}[1]{\iffalse #1 \fi}

\newcommand{\sname}{AFFT}
\newcommand{\lname}{Anticipative Feature Fusion Transformer}
\newcommand{\lsname}{\lname\ (\sname)}

\definecolor{Gray}{gray}{0.92}
\wacvfinalcopy % *** Uncomment this line for the final submission

\ifwacvfinal
\usepackage[breaklinks=true,bookmarks=false,colorlinks,citecolor=Green]{hyperref}
\else
\usepackage[pagebackref=true,breaklinks=true,colorlinks,bookmarks=false]{hyperref}
\fi

\begin{document}

%%%%%%%%% TITLE
\title{Anticipative Feature Fusion Transformer for Multi-Modal Action Anticipation}

\author[1,2\thanks{Equal contribution}]{Zeyun Zhong}
\author[2$^*$]{David Schneider}
\author[1]{Michael Voit}
\author[2]{Rainer Stiefelhagen}
\author[1,2]{Jürgen Beyerer}
\affil[1]{Fraunhofer IOSB, Karlsruhe \quad \texttt{\{firstname.lastname\}@iosb.fraunhofer.de}}
\affil[2]{Karlsruhe Institute of Technology (KIT) \quad \texttt{\{firstname.lastname\}@kit.edu}}

\maketitle
\thispagestyle{empty}

\input{sections/abstract}
\input{sections/introduction}
\input{sections/related_work}

\input{sections/method}
\input{sections/experiments}
\input{sections/results}
\input{sections/conclusion}

\textbf{Acknowledgements} This work was supported by the JuBot project which was made possible by funding from the Carl-Zeiss-Foundation. This work was performed on the HoreKa supercomputer funded by the Ministry of Science, Research and the Arts Baden-Württemberg and by the Federal Ministry of Education and Research.

{\small
\bibliographystyle{ieee_fullname}
\bibliography{egbib}
}

\clearpage
\input{sections/supplementary.tex}

\end{document}

%% file: sections/abstract.tex
%auto-ignore
\begin{abstract}
Although human action anticipation is a task which is inherently multi-modal, state-of-the-art methods on well known action anticipation datasets leverage this data by applying ensemble methods and averaging scores of uni-modal anticipation networks. In this work we introduce transformer based modality fusion techniques, which unify multi-modal data at an early stage. Our Anticipative Feature Fusion Transformer (AFFT) proves to be superior to popular score fusion approaches and presents state-of-the-art results outperforming previous methods on EpicKitchens-100 and EGTEA Gaze+. Our model is easily extensible and allows for adding new modalities without architectural changes. Consequently, we extracted audio features on EpicKitchens-100 which we add to the set of commonly used features in the community. \footnote{Code: \url{https://github.com/zeyun-zhong/AFFT}}
\end{abstract}

%% file: sections/introduction.tex
%auto-ignore
\section{Introduction}
\begin{figure}[t]
  \hspace*{-0.2cm}\includegraphics[width=\linewidth]{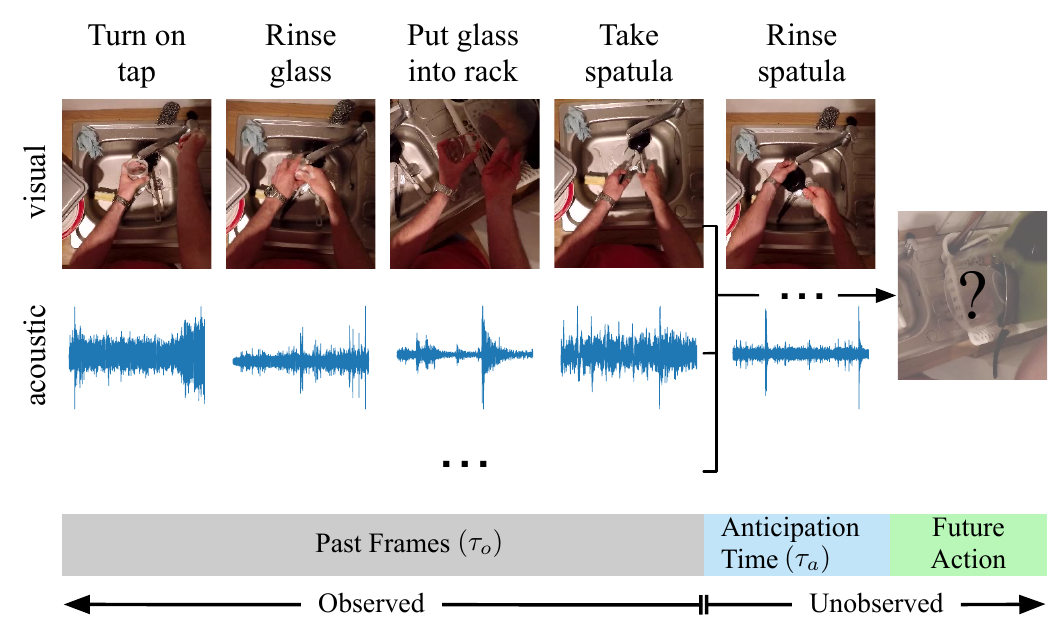}
  \caption{The action anticipation task aims to use the observed video segment of length $\tau_o$ to anticipate a future action $\tau_a$ seconds before it happens.}
  \label{fig:intro}
\end{figure}

Beyond human action recognition, anticipating possible future actions, as displayed in Figure~\ref{fig:intro}, is one of the most important tasks for human machine cooperation and robotic assistance, e.g. to offer a hand at the right time or to generate proactive dialog to provide more natural interactions. As the anticipation results are just assumptions, this tends to be significantly more challenging than traditional action recognition, which performs well with today’s well-honed discriminative models
~\cite{feichtenhofer2019slowfast,liu2022video}.
As modeling long temporal context is often crucial for anticipation~ \cite{furnariWhatWouldYou2019,senerTemporalAggregateRepresentations2020,girdharAnticipativeVideoTransformer2021}, many such methods were proposed in recent years, including clustering \cite{girdhar2017actionvlad,miechLearnablePoolingContext2017}, attention \cite{senerTemporalAggregateRepresentations2020} and recurrence \cite{furnariWhatWouldYou2019}.
While vision based systems are the de-facto standard for action anticipation \cite{furnariWhatWouldYou2019, girdharAnticipativeVideoTransformer2021, wuMeMViTMemoryAugmentedMultiscale2022}, additionally using other supporting modalities like optical flow features~\cite{wangTemporalSegmentNetworks2016,carreiraQuoVadisAction2017,kazakosEPICFusionAudioVisualTemporal2019} or knowledge about objects in the scene~\cite{furnari2017next} have shown to be beneficial.
In recent work \cite{kazakosEPICFusionAudioVisualTemporal2019,kazakosLittleHelpMy2021,nagraniAttentionBottlenecksMultimodal2021}, audio has been explored and shown to be complementary with appearance for action recognition in first-person vision. 
Consistent with most multi-modal action recognition models \cite{wangTemporalSegmentNetworks2016,carreiraQuoVadisAction2017}, anticipation models typically use score fusion (i.e., averaging predictions computed based on each single modality) to fuse different modalities. While averaging using fixed weights, including simple averaging \cite{senerTemporalAggregateRepresentations2020} and weighted averaging \cite{girdharAnticipativeVideoTransformer2021}, shows already superior results over the uni-modal baseline, Furnari et al.~\cite{furnariWhatWouldYou2019} show that assigning each modality with dynamical importance for the final prediction is particularly beneficial for anticipating egocentric actions. 
Inspired by the classical view of multisensory integration, i.e., information across the senses gets merged after the initial sensory processing is completed \cite{bloom1988brain,talsma2015predictive}, we take the mid-level fusion strategy in this work.
We present a transformer-based feature fusion model, \lsname, which successfully combines multi-modal features in a mid-level fusion process where features are first fused and the fused representations are utilized to anticipate next actions, different from all late and score fusion methods mentioned above. 
Our method is based on features and does not require end-to-end training of feature extractors. We see this as a major advantage since recent state-of-the-art results on various tasks have been driven by large foundation models which are difficult and resource intensive to train.
By combining strong feature extractors like \textsc{Omnivore}~\cite{girdharOmnivoreSingleModel2022} with mid-level feature fusion, we achieve state-of-the art results on common action anticipation datasets without the need for fine-tuning them.

In summary, our main contributions are: 
\begin{itemize}
    \item The \textbf{A}nticipative \textbf{F}eature \textbf{F}usion \textbf{T}ransformer (\sname), which successfully performs mid-level fusion on extracted features, improves significantly over score fusion based approaches and provides state-of-the-art performance on EpicKitchens-100 action anticipation and competing results on EGTEA Gaze+;
    \item A comparison of multiple self-attention and cross-attention based feature fusion strategies as well as detailed hyper parameter ablations for our final model;
    \item Extracted audio and \textsc{Omnivore}-based RGB features of EpicKitchen-100 which we provide to the community and an analysis of temporal and modality-wise performance contributions and model attention values.
\end{itemize}

%% file: sections/related_work.tex
%auto-ignore
\section{Related Work}
\textbf{Action anticipation} aims to predict future actions given a video clip of the past and present. While many approaches investigated different forms of action and activity anticipation from third person video \cite{gaoREDReinforcedEncoderDecoder2017,farhaWhenWillYou2018,keTimeConditionedActionAnticipation2019,gongFutureTransformerLongterm2022}, the first-person (egocentric) vision has recently gained popularity along with development of multiple challenge benchmarks to support it \cite{damen2018scaling,damen2020epic,li2018eye}. To model the temporal progression of past actions, \cite{furnariWhatWouldYou2019} proposed using an LSTM
to summarize the past and another LSTM for future prediction. \cite{senerTemporalAggregateRepresentations2020} made use of long-range past information and used an adapted version of the attention mechanism to aggregate short-term (`recent') and long-term (`spanning') features. To maintain the sequential temporal evolution while addressing the problem of modeling long-range temporal dependencies of recurrent architectures, a variation of GPT-2~\cite{radford2019language} has been recently proposed in \cite{girdharAnticipativeVideoTransformer2021}. We propose a transformer based feature fusion model to effectively fuse multiple modalities, and follow~\cite{girdharAnticipativeVideoTransformer2021} to use a generative language model for future action prediction.

\textbf{Multi-modal fusion for action anticipation.} The modalities typically used in prior work for egocentric vision are RGB, objects and optical flow \cite{furnariWhatWouldYou2019,senerTemporalAggregateRepresentations2020,wuLearningAnticipateEgocentric2021,zatsarynnaMultiModalTemporalConvolutional2021,girdharAnticipativeVideoTransformer2021}. To fuse information contained in different modalities, anticipation models typically utilize a late fusion strategy, similar to many multi-modal action recognition models \cite{wangTemporalSegmentNetworks2016,carreiraQuoVadisAction2017,kazakosEPICFusionAudioVisualTemporal2019}. These fusion methods can be broadly divided into score fusion and feature fusion. While in score fusion, the predicted future action scores of each modality are combined using either fixed weights, in form of simple averaging \cite{senerTemporalAggregateRepresentations2020,wuLearningAnticipateEgocentric2021} or weighted averaging \cite{girdharAnticipativeVideoTransformer2021}, or dynamic weights based on the scene \cite{furnariWhatWouldYou2019}, the feature fusion combines the predicted future action feature and an additional feed-forward layer is utilized to generate the action score \cite{zatsarynnaMultiModalTemporalConvolutional2021}. Different from the late fusion strategy, we take the mid-level fusion strategy inspired by the classical view of multisensory integration \cite{bloom1988brain,talsma2015predictive}. Specifically, we adopt the multi-head attention mechanism \cite{vaswaniAttentionAllYou2017} to combine different modalities at each timestamp and utilize the variation of GPT-2 following \cite{girdharAnticipativeVideoTransformer2021} to analyze the temporal evolution of the fused past features and predict future action features. Finally, a feed-forward layer is used to predict the future action class.

\textbf{Audio-visual learning.} Recent work used audio for an array of video understanding tasks, including self-supervised representation learning \cite{aytar2016soundnet,arandjelovicLookListenLearn2017a,korbarCooperativeLearningAudio2018}, audio-visual source separation \cite{owensAudioVisualSceneAnalysis2018a,afouras2018conversation,ephrat2018looking}, localizing sounds in video frames \cite{arandjelovic2018objects,senocak2019learning}, generating sounds from video \cite{owens2016visually,zhou2018visual,gao20192}, leveraging audio for efficient action recognition \cite{korbarSCSamplerSamplingSalient2019,gaoListenLookAction2020}, and utilizing audio to improve classification performance of action recognition \cite{kazakosEPICFusionAudioVisualTemporal2019,kazakosLittleHelpMy2021,nagraniAttentionBottlenecksMultimodal2021}. Different from all the work above, we focus on making use of audio as a complementary source of information for action anticipation.

%% file: sections/method.tex
%auto-ignore
\section{Methodology}
\begin{figure}[t]
  \includegraphics[width=\linewidth]{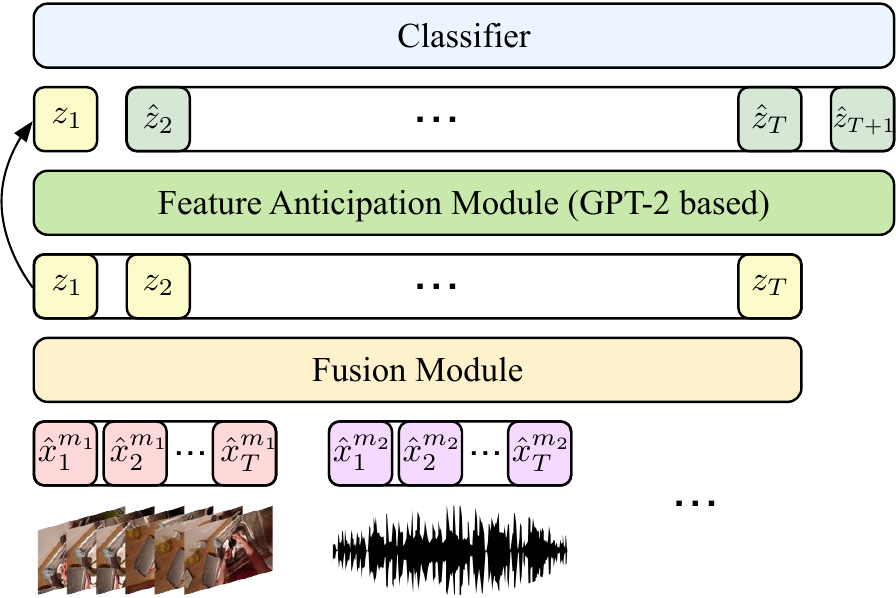}
  \caption{Architecture of \sname. The feature encoders are omitted, we directly list the feature vectors $\hat{x}^M$. A fusion module combines the modality specific feature vectors. The feature anticipation module then predicts the features of the next time step, followed by a linear classifier.}
\label{fig:arch}
\end{figure}

Our architecture which is displayed in Figure~\ref{fig:arch} consists of three exchangeable components: Modality specific feature extractors $f_\Pi^{m_j}, j\in\{1,\dots,M\}$, a cross-modal fusion module $g_\Phi$ and an anticipation module $a_\Omega$. Since this work analyzes multi-modal fusion on frozen features, we assume all $f_{\Pi}$ to have pretrained frozen weights and therefore refer to Section~\ref{sec:feature-extraction} for more details on the specific feature sets used for our experiments. Our proposed fusion modules are presented in Section~\ref{sec:fusion-module}. We follow \cite{girdharAnticipativeVideoTransformer2021} and use a variation of the GPT-2~\cite{radford2019language} model as feature anticipation module to predict $\hat{z}_{i+1} = a_\Omega(z_i), i\in\{1,\dots,T\}$.

\subsection{Problem statement}
In this work, we follow the anticipation setup defined in \cite{damen2018scaling,damen2020epic}. As illustrated in Figure \ref{fig:intro}, the action anticipation task aims to predict an action starting at time $\tau_s$ by observing a video segment of length $\tau_o$. The observation segment is $\tau_a$ seconds preceding the action, i.e., from time $\tau_s$ $-$ $(\tau_a \, + \, \tau_o )$ to $\tau_s$ $-$ $\tau_a$,  where $\tau_a$ denotes the ``anticipation time'', i.e., how many seconds in advance actions are to be anticipated. The anticipation time $\tau_a$ is usually fixed for each dataset, whereas the length of the observation segment is typically dependent on the individual method. In our experiments we assume $T$ temporally sequential input  observations $x_i^{m_j}, i \in \{1,\dots,T\}, j \in \{1,\dots,M\}$ which describe the observation time $\tau_o$ for each of the $M$ available modalities. The anticipated action is defined to be at time step $T+1$ without observation and label $y_{T+1}$. Depending on the dataset, the preceding observations might additionally be labelled with $y_{i}$. Since this work is aimed at feature based modality fusion, we assume fixed feature extractors $f_\Pi$ and define the individual extracted features as $\hat{x}_i^{m_j} = f^{m_j}_\Pi(x_i^{m_j})$ and the collection of all $T\times M$ features for an input sample as $\hat{x}^M$. 

\subsection{Cross-modal fusion}
\label{sec:fusion-module}

\begin{figure*}[t]
\centering
  \includegraphics[width=0.9\textwidth]{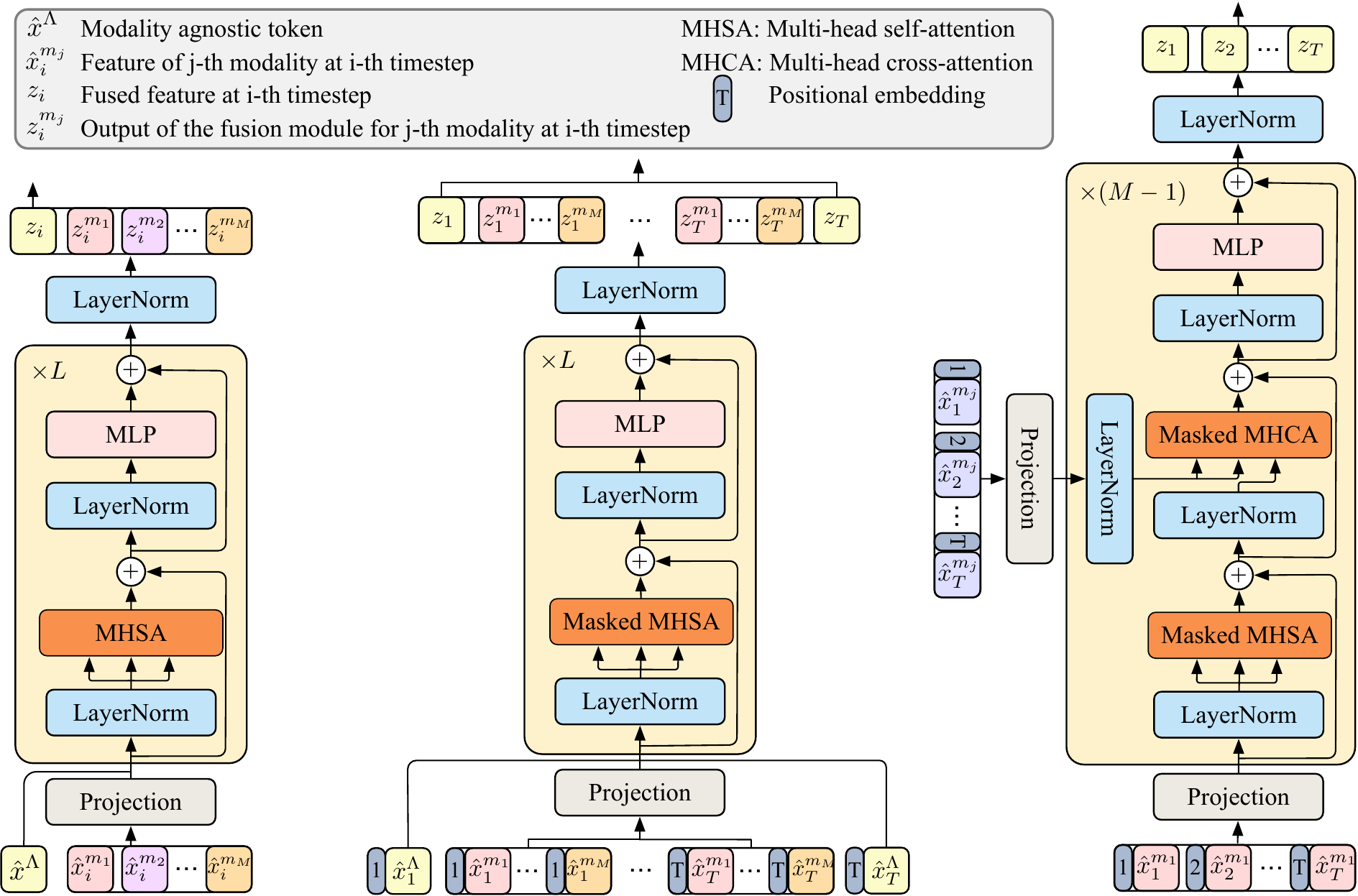}
  \caption{The \emph{SA-Fuser} on the left applies Transformer Encoder blocks at individual time steps while the \emph{T-SA-Fuser} in the middle and the transformer decoder based \emph{CA-Fuser} on the right perform fusion on the whole temporal sequence at once.}
  \label{fig:fusion-individual}
\end{figure*}

\paragraph{Time-decoupled feature fusion}
\label{sec:cmtfuser}
In order to fuse the features $\hat{x}_i^M$ on each individual time-step  separately, we apply $L$ consecutive transformer encoder blocks as used in \cite{dosovitskiyImageWorth16x162021} with dimensionality $d$ and $k$ attention heads, this module is displayed on the left of Figure~\ref{fig:fusion-individual}.  We found that modality-wise positional embeddings do not yield an improvement of performance, presumably since the modality specific features are already easily separable in feature space. We do ablate the usage of a modality agnostic learnable token $\hat{x}^\Lambda$, similar to the concept of a learnable class-token used in \cite{dosovitskiyImageWorth16x162021}. The module with the prepended learnable token $x^{\Lambda}$ is referred to as \emph{Self-Attention Fuser (SA-Fuser)}. Without this learnable token we average the resulting output tokens $z_i^{m_j}$. We consider the usage of the learnable token as default, experiments without token are marked as such.

\paragraph{Temporal feature fusion}
\label{sec:t-cmtfuser}
The \emph{Temporal Self-Attention Fuser (T-SA-Fuser)} which is displayed in the middle of Figure~\ref{fig:fusion-individual} follows the paradigm of the \emph{SA-Fuser}, but instead of fusing multi-modal features per time step, all modality features for all time steps are used to provide all output features $z = g(\hat{x}^M)$ at once. A learnable positional embedding $p_i$ is used to encode the temporal positions for each modality and an attention mask enforces that an output feature at temporal position $i$ only attends to previous or concurrent multi-modal features. Instead of a single modality agnostic token, we provide a learned token for each time step provided to the module. Learned positional embeddings are added to the tokens of each time step to allow the model to differentiate them.
\paragraph{Temporal cross-attention feature fusion}
Our third fusion module, which is displayed on the right of Figure~\ref{fig:fusion-individual}, is inspired by \cite{huang2020multimodal} and follows a different paradigm. Instead of providing all modalities at once, we iteratively enrich a main modality (RGB in our experiments) with information from other modalities. Instead of $L$ transformer encoder blocks, $(M-1)$ transformer decoder blocks \cite{vaswaniAttentionAllYou2017} are used. Following the decoder architecture, the RGB features $\hat{x}^{RGB}$ are provided as main input which provides the queries for the multi-head cross-attention and each block makes use of another modality $\hat{x}^{m_j}$ as second decoder input which provides the keys and values. Positional embeddings are added to all modality features. We do not use additional tokens, but rather directly predict the fused features $z$. We refer to this module as \emph{Cross-Attention Fuser (CA-Fuser)}

\subsection{Feature anticipation and classification}
\label{sec:classification}
After different modality features get fused by the fusion module, a variation of the GPT-2~\cite{radford2019language} model is used to predict the future features $\hat{z}_{i+1} = a_\Omega(z_i), i\in\{1,\dots,T\}$, following~\cite{girdharAnticipativeVideoTransformer2021}. To encode the temporal ordering and obtain generative ability, learnable positional embeddings and a temporal attention mask are used.
Based on the anticipated features $\hat{z}$ we define a classification head $h$, %which is used for feature based action classification. $h$ is
a single linear layer followed by a softmax activation function. The anticipation result is based on the predicted future feature, so $\hat{y}_i = h(\hat{z}_{i})$ and the final anticipation result $\hat{y}_{T+1} = h(\hat{z}_{T+1})$.

\subsection{Loss functions}
\label{sec:losses}
Our loss functions follow the setting of \cite{girdharAnticipativeVideoTransformer2021}. We apply three losses $\mathcal{L} = \mathcal{L}_\mathrm{next} + \mathcal{L}_\mathrm{cls} + \mathcal{L}_\mathrm{feat}$.
$\mathcal{L}_\mathrm{next}$ is defined on $\hat{y}_{T+1}$ and $y_{T+1}$ according to the task of action anticipation. Since the network output does not only provide features $\hat{z}_{T+1}$ for the anticipated next action but also for the preceding time steps $i \in \{1,\dots,T\}$,   $\mathcal{L}_\mathrm{cls}$ evaluates the action classification performance of these preceding features, so on $\hat{y}_{i}=h(\hat{z}_i)$ and $y_{i}$. Both are cross-entropy losses. $\mathcal{L}_\mathrm{feat}$ is the mean squared error between predicted and fused features $\hat{z}_{i}$ and $z_i$.

%% file: sections/experiments.tex
%auto-ignore
\section{Experimental Setup}
In order to investigate the influence of the different fusion strategies and evaluate the proposed method for the action anticipation task, we train and evaluate our methods on two different datasets (discussed in detail in Section~\ref{sec:datasets}). To allow a fair comparison with prior work, we first use pre-extracted TSN features~\cite{wangTemporalSegmentNetworks2016} as input features for both datasets provided by~\cite{furnariWhatWouldYou2019}. To investigate the impact of the audio modality for action anticipation, we train a TSN audio action recognition model following~\cite{kazakosEPICFusionAudioVisualTemporal2019} and extract its features for fusion with other modalities. In order to show the generalization of our proposed fusion method, we extract alternative RGB features from a recent state-of-the-art visual model, \textsc{Omnivore}~\cite{girdharOmnivoreSingleModel2022}. Information regarding feature extraction is discussed in detail in Section~\ref{sec:feature-extraction}. All experiments follow the training procedure described in Section~\ref{sec:implementation}.

\subsection{Datasets and metrics}
\label{sec:datasets}
We perform experiments on two large-scale egocentric (first-person) video datasets: EpicKitchens-100~\cite{damen2020epic} and EGTEA Gaze+~\cite{li2018eye}. EpicKitchens-100 consists of 700 long unscripted videos of cooking activities totalling 100 hours. It contains 90.0K action annotations, 97 verbs, and 300 nouns. We considered all unique (\textit{verb}, \textit{noun}) pairs in the public training set, obtaining 3,807 unique actions. We use the official train, val and test splits to report performance. The test evaluation is performed on a held-out set through a submission to the official challenge server. EGTEA Gaze+ is another popular egocentric action anticipation dataset. It contains 10.3K action annotations, 19 verbs, 51 nouns and 106 unique actions.

We report class mean top-5 recall~\cite{furnariLeveragingUncertaintyRethink2018} for EpicKitchens-100, a class-aware metric in which performance indicators obtained for each class are averaged to obtain the final score, accounting for the multi-modality in future predictions and class imbalance in a long-tail distribution. For EGTEA Gaze+, we report top-1/5 and class mean top-1. As some prior works report their results averaged across the official three splits, and some evaluate their methods on the first split only, we test our method using both recipes.% In all tables, the metric used to rank methods in the official challenge leaderboards corresponds to future action (act.) prediction.

\subsection{Uni-modal features}
\label{sec:feature-extraction}
\noindent\textbf{RGB.} We compare two types of RGB features, the commonly used TSN features~\cite{wangTemporalSegmentNetworks2016} provided by~\cite{furnariWhatWouldYou2019} and Swin transformer~\cite{liu2021swin} features which we extracted with \textsc{Omnivore}~\cite{girdharOmnivoreSingleModel2022} to represent more recent transformer based approaches. Both feature extractors are trained for action recognition. While TSN features are extracted by applying TSN on each frame, we extract Swin features by feeding 32 consecutive past frames totalling 1.067s video with a frame rate of 30fps to the \textsc{Omnivore} model for each timestamp.

\noindent\textbf{Audio.} Following \cite{kazakosEPICFusionAudioVisualTemporal2019}, We extract 1.28s of audio, convert it to single-channel, and resample it to 24kHz. We then convert it to a log-spectrogram representation using an STFT of window length 10ms, hop length 5ms and 256 frequency bands, resulting in a 2D spectrogram matrix of size 256$\times$256, after which we compute the logarithm. Different from~\cite{kazakosEPICFusionAudioVisualTemporal2019}, we extract audio in an online manner, i.e., we extract the past audio segment for each timestamp, prohibiting the model to have access to the future, which is the prerequisite for the anticipation task. We feed such matrices to the TSN network, train it for the action recognition task and extract features for our work.

\noindent\textbf{Objects and optical flow.} We use the existing object and optical flow features provided by~\cite{furnariWhatWouldYou2019}. Object representations are obtained by accumulating the confidence scores of all bounding boxes predicted by a Faster R-CNN~\cite{ren2015faster} for each object class. Optical flow features are extracted by feeding 5 consecutive past frames of horizontal and vertical flow, forming a tensor with 10 channels, to a TSN model trained for action recognition.

\subsection{Implementation details}
\label{sec:implementation}
\noindent\textbf{Architecture details.} For our \sname{} model we use the marked default hyper parameters from Table~\ref{tab:fuser_architecture}. For EGTEA Gaze+, we reduce the number of layers of the fuser and the future predictor to 2, since EGTEA Gaze+ is relatively small compared to EpicKitchens-100. We employ a linear projection layer for modality features that are not in alignment with the hidden size of the fuser. To match the hidden dimension used in the future predictor, another linear layer is employed to project the fused modality features.

\noindent\textbf{Training \& testing.}
We sample all modality features at 1 fps, resulting in a sequence of feature vectors whose length corresponds to observation time $\tau_o$. Default observation time is 10s, the other observation lengths are analyzed in Section~\ref{sec:temporal-context}. We train our models with SGD+momentum using $10^{-6}$ weight decay and $10^{-3}$ learning rate for 50 epochs, with 20 epochs warmup~\cite{goyal2017accurate} and 30 epochs of cosine annealed decay, following~\cite{girdharAnticipativeVideoTransformer2021}. We use mixup data augmentation~\cite{zhang2017mixup} with $\alpha$ $=$ 0.1. Default settings for dropout and the stochastic depth regularization technique~\cite{huang2016deep} are listed in Table~\ref{tab:fuser_architecture}. Following standard practice~\cite{furnariWhatWouldYou2019,girdharAnticipativeVideoTransformer2021,wuMeMViTMemoryAugmentedMultiscale2022}, our model is optimized to predict the action label during training and marginalize the output probabilities to obtain the verb and noun predictions in testing.

%% file: sections/results.tex
%auto-ignore
\section{Results}
\input{tables/ablation_fusion_strategies.tex}
In Section~\ref{sec:ablation-fusion} we ablate the proposed fusion architectures. Continuing with the best architecture, we find optimized hyper parameters in Section~\ref{sec:ablation:fuserparams} and the optimal temporal context in Section~\ref{sec:temporal-context}. In Section~\ref{sec:ablation-modalities} we analyze the contribution of individual modalities to the final model performance and in Section~\ref{subsec:sota} our models are compared against state-of-the-art feature based action anticipation models on EpicKitchens-100 and EGTEA Gaze+. The models trained with RGB-TSN and RGB-Swin features are referred to \sname-TSN and \sname-Swin respectively.

\subsection{Fusion strategies}
\label{sec:ablation-fusion}
We evaluate the fusion architectures presented in Section~\ref{sec:fusion-module} against score fusion based methods and evaluate which of our strategies proves best for multi-modal fusion. Table~\ref{tab:comparison_fusion} lists all methods. In our comparison we include \emph{Modality Attention (MATT)}~\cite{furnariWhatWouldYou2019}, a learned score fusion weighting method, but find it to be lacking in our setting. 
For score averaging and weighted averaging, we choose the same setting as~\cite{girdharAnticipativeVideoTransformer2021}, verifying their results.
Combining temporal and modality attention as done with \emph{T-SA-Fuser} performs worst in our feature fusion models, which we assume to be caused by the complexity of this process. \emph{CA-Fuser} introduces an inductive bias by introducing a new modality with each consecutive block, splitting the process of attention into separate smaller problems instead of presenting all temporal and modality tokens at once. Our best approach \emph{SA-Fuser} on the other hand is even simpler, since it splits the problem along time-steps and only attends over the modality tokens. Temporal attention is then performed in a completely separate step with the GPT-2 based future predictor. We believe this reduced complexity to be the mechanism which leads to optimal performance of our final model.
For further experiments we use the \emph{SA-Fuser} as our default fusion module.

\input{tables/ablation_cmtfuser_params.tex}

\subsection{Architecture ablations}
\label{sec:ablation:fuserparams}
In Table~\ref{tab:fuser_architecture}, we ablate different hyper parameters of our architecture. The default parameters are marked with grey table cells, the best values are typed in boldface.

\noindent\textbf{Projection layer and common dimensionality.} The dimension of all multi-modal input features must coincide. This could be achieved using a simple linear layer, a linear layer with ReLU activation function \cite{kazakosLittleHelpMy2021,gongFutureTransformerLongterm2022} or a gated linear projection \cite{miechLearnablePoolingContext2017,shvetsovaEverythingOnceMultiModal2022}, listed in Table~\ref{tab:sub_mapping}. We add an additional variant \textit{sparse linear}, meaning a linear layer is only applied for features which have a different dimension than the desired common dimension and show that it outperforms other projection methods. In Table~\ref{tab:sub_dim}, we examine how the projection dimension influences performance. We find a dimensionality of 1024 to be optimal, a higher dimension presumably decreases performance due to the increased number of parameters and overfitting effects.

\noindent\textbf{Attention heads and encoder blocks.} We compare the impact of different head numbers of the encoder multi-head attention in Table~\ref{tab:sub_heads}, the number of encoder blocks is analyzed in Table~\ref{tab:sub_layers}, we find eight heads and six consecutive encoder blocks to be best.

\noindent\textbf{Effect of regularization.} We ablate using either no dropout and no stochastic depth~\cite{huang2016deep} (i.e. no regularization) or using stochastic depth with maximal layer dropping probability of 0.1. Results in Table~\ref{tab:sub_regularization} show that both dropout and stochastic depth regularization are very beneficial.

\subsection{Impact of temporal context.} 
\label{sec:temporal-context}

\begin{figure}
\centering
\begin{minipage}[t]{.5\linewidth}
  \centering
  \includegraphics[width=\linewidth,valign=t]{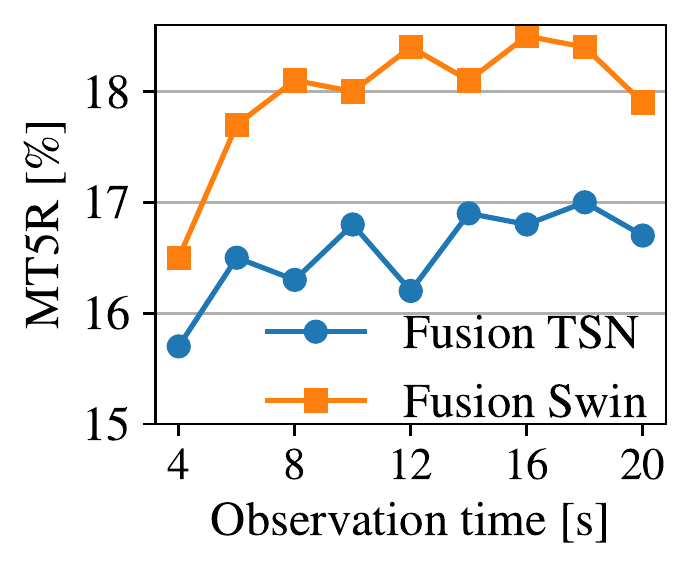}
  \captionof{figure}{Impact of temporal context on the validation set of EpicKitchens-100. Our method leverages long-term dependencies to improve anticipation performance.}
  \label{fig:temporal_context}
\end{minipage}\hspace{0.15cm}
\begin{minipage}[t]{.45\linewidth}
  \centering
  \includegraphics[width=\linewidth,valign=t]{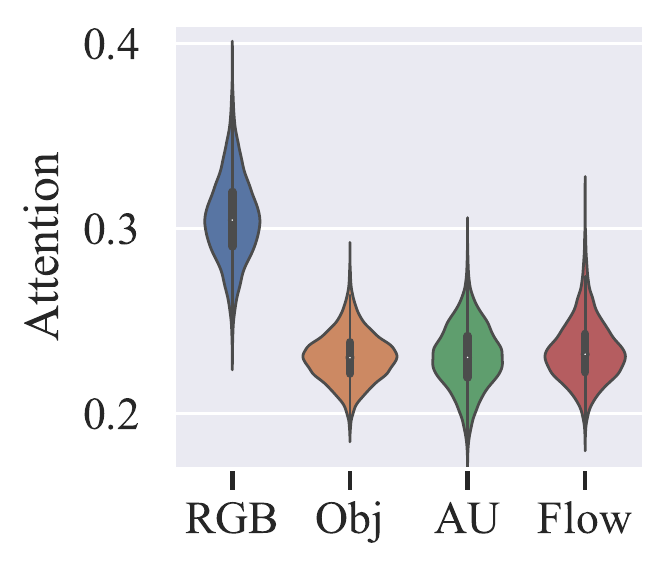}
  \vspace{0.24cm}
  \captionof{figure}{Modality attentions of \sname-Swin on the validation set of EK-100. Our method learns to pay more attention to RGB without any supervision.}
  \label{fig:modality_attention_swin}
\end{minipage}
\end{figure}

\begin{figure}[t]
    \centering
    \includegraphics[width=0.9\linewidth]{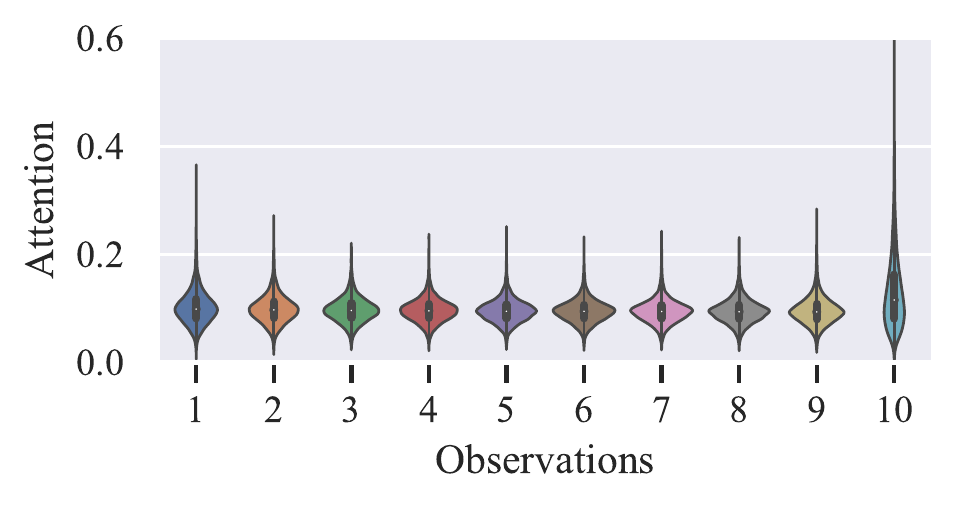}
    \caption{Temporal attentions of \sname-Swin over all samples of the validation set of EK-100. Our method attends not only to the recent past, but also to the entire past frames.}
    \label{fig:temporal_attention_swin}
\end{figure}
\begin{figure*}[t]
    \centering
    \begin{subfigure}{\linewidth}
        \includegraphics[width=\linewidth]{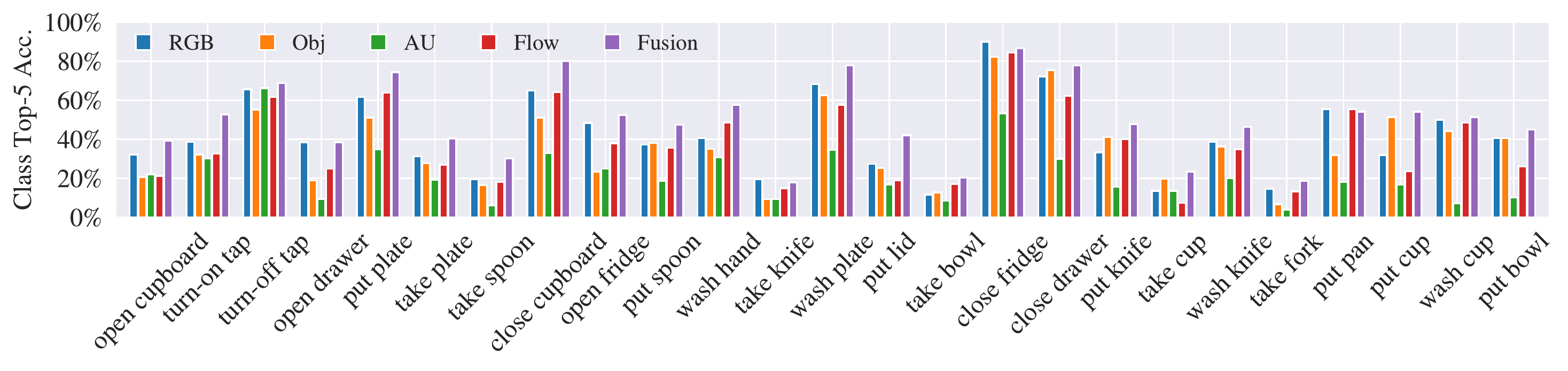}
    \end{subfigure}
    \vspace{-0.6cm}
    \caption{Per-class top-5 accuracy of fusion (\sname-TSN) and single modalities for the largest 25 actions in the validation set of EpicKitchens-100. The classes are presented in the order of number of samples per class, from left to right. For most classes the fusion method provides significantly better results over the single modalities.}
    \label{fig:class_acc_tsn}
\end{figure*}
\begin{figure}[t]
   \centering
\includegraphics[width=\linewidth]{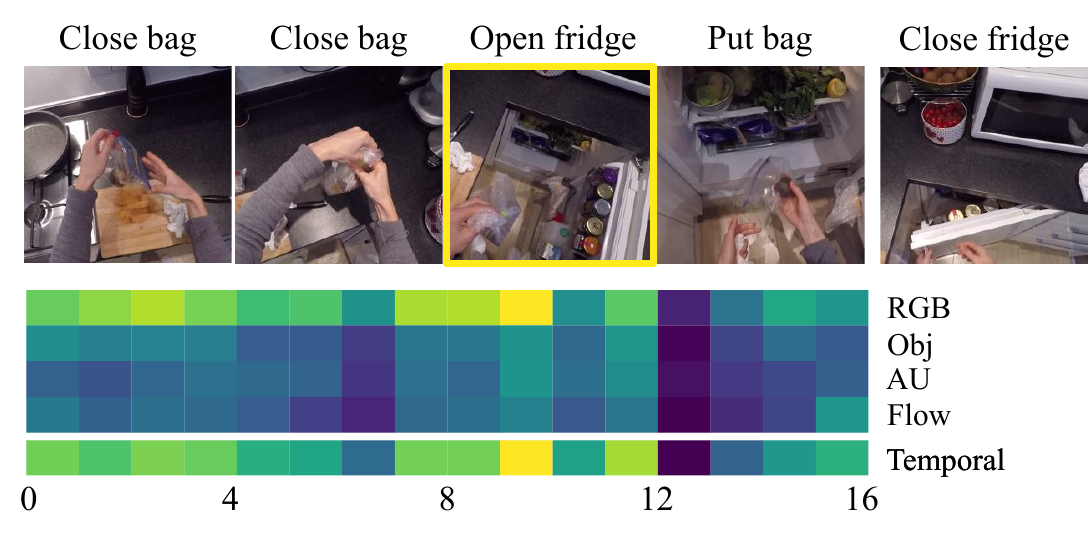}
     \caption{Qualitative results on EpicKitchens-100. The horizontal and vertical axes indicate the index of past frames as well as the modality. The closer the color is to yellow, the higher the attention score. A video frame is highlighted  with a yellow box when the attention score of the frame is highly activated.}
     \label{fig:fridge-attention}
\end{figure}

To study the ability of modeling sequences of long-range temporal interactions, we train and test the model with different lengths of temporal context, i.e., observation time $\tau_o$. As seen in Figure \ref{fig:temporal_context}, as more frames of context are incorporated, the performance improves for both, \sname-TSN and \sname-Swin. The gains are especially pronounced when trained using RGB-Swin features (16.5 $\rightarrow$ 18.5 $=$ 2.0 $\uparrow$) vs. RGB-TSN features (15.7 $\rightarrow$ 17.0 $=$ 1.3 $\uparrow$). To further explore how the temporal context is utilized, following \cite{girdharAnticipativeVideoTransformer2021}, we extract temporal attentions from the last layer of the feature anticipation module for all samples in the validation set of EpicKitchens-100, average them over heads and visualize them in Figure \ref{fig:temporal_attention_swin}. The anticipation module learns to attend to visual features in the recent past, showing that the nearest past frames provide crucial keys for predicting future actions. This aligns with previous work \cite{keTimeConditionedActionAnticipation2019,senerTemporalAggregateRepresentations2020} which reflects the importance of the recent past in designing anticipation models. However, while the median attention values of more distant past frames are smaller (close to 0.1), the attention distribution is significantly scattered, indicating that the model can choose to attend to important actions not only from the recent past, but also from the entire observation time, as illustrated in an example in Figure~\ref{fig:fridge-attention}. Here the model attends to an early time step in the middle of the observation which shows the opening of a fridge in order to predict the the future action `close fridge'. Results for \sname-TSN are listed in the supplementary.

\subsection{Modality contributions}
\label{sec:ablation-modalities}

As shown in Table \ref{table:sub_single}, visual modalities, especially RGB, have higher performances than audio, also observable in Figure \ref{fig:class_acc_tsn}. Benefiting from a larger model capacity and better representative ability, RGB features extracted with an Omnivore-pre-trained Swin-Transformer perform significantly better than TSN features.
Results in Table \ref{tab:sub_fusion} show that the anticipation performance keeps increasing when additional modalities are introduced for both kinds of RGB features. In particular, \sname-TSN and \sname-Swin have gains of 3.6\% and 1.9\% over their uni-modal RGB performances in Table~\ref{table:sub_single}, respectively.
Per-class top-5 accuracies, for individual modalities as well as for our fusion model (\sname-TSN) trained on all four modalities, can be seen in Figure \ref{fig:class_acc_tsn}. The fusion model outperforms uni-modal models for most classes, often by a significant margin. Results of \sname-Swin are shown in the supplementary material.
To analyze the contribution of our extracted audio features, we conduct experiments with the visual modalities (RGB+Obj+Flow) only and compare them with models trained on all four modalities, which results in an increase of 0.6\% (\sname-TSN) and 0.4\% (\sname-Swin) in mean top-5 action anticipation accuracy as seen in Table \ref{tab:sub_fusion}. To further validate the benefit of audio, we compute a confusion matrix with the utilization of audio for the largest-15 action classes, following \cite{kazakosEPICFusionAudioVisualTemporal2019}, which we list in the supplementary.
To better understand how the fusion module models relative importance of different modalities, we visualize the learned modality attentions of \sname-Swin in Figure \ref{fig:modality_attention_swin}. Specifically, we use attention rollout~\cite{abnar2020quantifying} to aggregate attention over heads and layers. As shown in the figure, RGB has gained the most attention, indicating the modality which contributes the most for the anticipation task (as seen in Table \ref{table:sub_single}) will be automatically utilized most by the fusion module, as would be expected. Figure \ref{fig:modality_attention_swin} also shows that the attention distributions of all modalities spread widely, showing that the model learns to adjust the relative importance of individual modalities based on each sample.

\input{tables/ablation_modalities_new}

\subsection{Comparison to the state-of-the-art}
\label{subsec:sota}
\input{tables/results_ek100.tex}

Our final models follow the default hyper parameters from Table~\ref{tab:fuser_architecture}. On EpicKitchens-100, \sname-TSN and \sname-Swin use observations of 18s and 16s respectively, while the default observation time (10s) is used for EGTEA Gaze+. For the comparisons, we distinguish between training with frozen backbones (i.e., training on frozen features) and training with fine-tuned backbones (marked with gray font). In all tables in this section, the main metrics used to rank methods for these datasets are highlighted.

In Table \ref{tab:ek100_sota}, we compare our method with state-of-the-art methods on EpicKitchens-100. The table is divided into two compartments according to the validation and test splits. On the validation split, our \sname-TSN outperforms other fusion methods with a large margin (14.8 $\rightarrow$ 16.4 $=$ 1.6 $\uparrow$) with the exact same features provided by~\cite{furnariWhatWouldYou2019}. With the addition of audio, the performance is further improved by 0.6\%. \sname-Swin$^+$ which uses Omnivore features outperforms the current state-of-the-art model MeMViT by 0.8\% mean top-5 ratio action anticipation performance on the val split without the need to fine-tune the backbone network.
Consistent with the results on validation split, our method also outperforms prior fusion methods on the test set of EpicKitchens-100. As shown in bottom compartment in Table \ref{tab:ek100_sota}, we get the largest gains on tail classes, for which our method proves particularly effective. Note that Table~\ref{tab:ek100_sota} lists peer-reviewed results, only. In our supplementary we also list results of the EpicKitchens-Challenge, which holds many non-peer-reviewed results, often created with model ensembling of various methods.

Next we evaluate our method on EGTEA Gaze+, shown in Table \ref{tab:egtea_sota}. Following prior works \cite{liuForecastingHumanObjectInteraction2020,girdharAnticipativeVideoTransformer2021}, we set the anticipation time $\tau_a$ to 0.5s. As some prior works report the results averaged across the three official splits, while others test on split 1 only, we evaluate our methods using both recipes. Using fixed features, \sname-TSN outperforms prior works using both recipes, especially for class mean top-1.

\input{tables/results_egtea.tex}

%% file: tables/ablation_fusion_strategies.tex
\begin{table}[t]
\centering
\begin{tabular}{rlc} 
\toprule\ & Fusion strategy & Act.\\ 
\midrule
\multirow{3}{*}{\rotatebox{90}{Score}} & Average  & 16.4 \\
    & Weighted average  & 17.3 \\
    & MATT & 12.2 \\ 

\\[-.9em]
\arrayrulecolor{gray}
\hhline{*{1}{~}*{2}{-}}
\arrayrulecolor{black}
\\[-.9em]

\multirow{5}{*}{\rotatebox{90}{Feature}} 
    & SA-Fuser (w/o Token) & 17.1     \\
    & \textbf{SA-Fuser} & \textbf{18.0}  \\
    & T-SA-Fuser & 15.2 \\
    & CA-Fuser & 16.6   \\
\bottomrule
\end{tabular}
\caption{Comparison of fusion strategies. The results are based on all modalities with RGB-Swin features. We refer to the model using the \emph{SA-Fuser} as our method \sname.}
\label{tab:comparison_fusion}
\end{table}

%% file: tables/ablation_cmtfuser_params.tex
\begin{table}[t]
    \centering
    \begin{subtable}[t]{.38\linewidth}
    \resizebox{\linewidth}{!}{
    \begin{tabular}[t]{l|c}
         Projection & Act. \\ \hline
         Lin. & 17.6\\
         \cellcolor{Gray}Lin. (sparse) & \cellcolor{Gray}\textbf{18.0} \\
         Lin. + ReLU & 17.1\\
         GLU \cite{miechLearnablePoolingContext2017} & 17.8\\
    \end{tabular}}
    \vspace{0.32cm}\caption{Projection layer.}
    \label{tab:sub_mapping}
    \end{subtable}
    \hfill
    \begin{subtable}[t]{.25\linewidth}
    \resizebox{\linewidth}{!}{
    \begin{tabular}[t]{c|c}
         Dim. & Act. \\ \hline
         512 &  16.7\\
         768 & 17.2 \\
         \cellcolor{Gray}1024 & \cellcolor{Gray}\textbf{18.0} \\
         1280 & 18.0\\
         %1536 & \textbf{18.1}\\
         2048 & 16.9\\
    \end{tabular}}
    \caption{Dimension.}
    \label{tab:sub_dim}
    \end{subtable}
    \hfill
    \begin{subtable}[t]{.27\linewidth}
    \resizebox{\linewidth}{!}{
    \begin{tabular}[t]{c|c}
         Heads & Act. \\ \hline
         \cellcolor{Gray}4 & \cellcolor{Gray}18.0\\
         8 & \textbf{18.4}\\
         16 & 17.4\\
    \end{tabular}}
    \vspace{0.72cm}\caption{No. of heads.}
    \label{tab:sub_heads}
    \end{subtable}
    \begin{subtable}[t]{.28\linewidth}
    \resizebox{\linewidth}{!}{
    \begin{tabular}[t]{c|c}
         Layers & Act. \\ \hline
         2 & 17.9 \\
         4 & 17.3\\
         \cellcolor{Gray}6 & \cellcolor{Gray}\textbf{18.0}\\
    \end{tabular}}
    \caption{No. of layers.}
    \label{tab:sub_layers}
    \end{subtable}
    \begin{subtable}[t]{.58\linewidth}
    \resizebox{\linewidth}{!}{
    \begin{tabular}[t]{l|c}
         Regularization & Act. \\ \hline
         no regularization &  15.8\\
         stochastic depth (0.1) & 16.2 \\
         \cellcolor{Gray}drop. \& stoch. depth (0.1) & \cellcolor{Gray}\textbf{18.0}\\
    \end{tabular}}
    \caption{Regularization.}
    \label{tab:sub_regularization}
    \end{subtable}
    \caption{Fuser architecture ablation on the validation set of EpicKitchens-100. 
    %If not specified, the default setting is: the fuser has depth 6, width 1024 and 4 heads. 
    Default settings are marked in \colorbox{Gray}{gray}.}
    \label{tab:fuser_architecture}
\end{table}

%% file: tables/ablation_modalities_new.tex
\begin{table}[t]
	\centering
	\setlength\tabcolsep{2pt}
	\begin{subtable}[t]{0.4\linewidth}
 \centering
            
			\begin{tabular}[t]{llc}
				\toprule
				Mod. & Backbone & Act.\\
				\midrule
				RGB & TSN & 13.2 \\
				RGB  & Swin & 16.1 \\
				Obj  & F. R-CNN &9.9 \\
				AU  & TSN& 5.3 \\
				Flow  & TSN & 7.5 \\ 
				\bottomrule	
		\end{tabular}
	\vspace{0.7cm}
	\caption{Results of individual modalities.}
	\label{table:sub_single}
	\end{subtable}\hfill
	\begin{subtable}[t]{0.59\linewidth}
 \centering
            
			\begin{tabular}[t]{lcc}
				\toprule
			\multicolumn{1}{l|}{RGB}	& TSN & Swin  \\ 
				\arrayrulecolor{gray}
                    \cmidrule(lr){2-2} 
                    \cmidrule(lr){3-3}
                    \arrayrulecolor{black} 
                    \\[-.9em]
				 \multicolumn{1}{l|}{Other} & Act. & Act. \\
				\cmidrule{1-3}
				\multicolumn{1}{l|}{Obj} & 15.9 & 16.7  \\
				\multicolumn{1}{l|}{AU} & 15.4 & 16.8  \\
				\multicolumn{1}{l|}{Flow} & 15.2 & 16.5  \\
				\multicolumn{1}{l|}{Obj+Flow} & 16.2 & 17.6 \\
				\multicolumn{1}{l|}{Obj+AU+Flow} & 16.8 & 18.0 \\
				\bottomrule
		\end{tabular}
		\caption{Results of multiple modalities combined with RGB.}
		\label{tab:sub_fusion}
	\end{subtable}
	\caption{Impact of individual modalities on the validation set of EpicKitchens-100. Compared to other modalities, RGB performs significantly better, particularly on features extracted by Swin. The proposed fusion method benefits from multi-modal inputs. The more modalities are provided, the better the anticipation model performs.}
	\label{tab:impact_modality}
\end{table}

%% file: tables/results_ek100.tex
\begin{table}[t]
    \centering
    \setlength\tabcolsep{1.5pt}
        \resizebox{\linewidth}{!}{
    \begin{tabular}{clcc>{\columncolor{Gray}}c>{\enspace}ccc>{\enspace}ccc}
        \toprule
         \multirow{2}{*}{} & \multirow{2}{*}{Method} & \multicolumn{3}{c}{Overall} & \multicolumn{3}{c}{Unseen Kitchen} & \multicolumn{3}{c}{Tail Classes}\\ 
         \arrayrulecolor{gray} \cmidrule(lr){3-5} \cmidrule(lr){6-8} \cmidrule(lr){9-11} \arrayrulecolor{black}
          &  & Verb & Noun & Act. & Verb & Noun & Act. & Verb & Noun & Act. \\ \midrule
         
         %------------Validation set--------
         \multirow{7}{*}{\rotatebox{90}{Val}} & chance & 6.4 & 2.0 & 0.2 & 14.4 & 2.9 & 0.5 & 1.6 & 0.2 & 0.1 \\
         & \color{gray}AVT+ \cite{girdharAnticipativeVideoTransformer2021} & \color{gray}28.2 & \color{gray}32.0 & \color{gray}15.9 & \color{gray}29.5 & \color{gray}23.9 & \color{gray}11.9 & \color{gray}21.1 & \color{gray}25.8 & \color{gray}14.1 \\
         & \color{gray}MeMViT \cite{wuMeMViTMemoryAugmentedMultiscale2022} & \color{gray}32.2 & \color{gray}37.0 & \color{gray}17.7 & \color{gray}28.6 & \color{gray}27.4 & \color{gray}15.2 & \color{gray}25.3 & \color{gray}31.0 & \color{gray}15.5 \\
         
         \arrayrulecolor{lightgray}\hhline{*{1}{~}*{10}{-}}\arrayrulecolor{black} \rule{0pt}{2.6ex}
         
         & RULSTM \cite{furnariWhatWouldYou2019} & \textbf{27.8} & 30.8 & 14.0 & \textbf{28.8} & \textbf{27.2} & 14.2 & \textbf{19.8} & 22.0 & 11.1 \\
         & TempAgg \cite{senerTemporalAggregateRepresentations2020} & 23.2 & 31.4 & 14.7 & 28.0 & 26.2 & 14.5 & 14.5 & 22.5 & 11.8 \\
         & AVT+-TSN \cite{girdharAnticipativeVideoTransformer2021} & 25.5 & 31.8 & 14.8 & 25.5 & 23.6 & 11.5 & 18.5 & 25.8 & 12.6 \\

         & \emph{Ours-TSN} & 21.3&32.7 & 16.4 &24.1 &25.5 &13.6 &13.2 &25.8 &14.3 \\
         & \emph{Ours-TSN$^+$} & 22.3 & 31.5 & 17.0 & 23.8 & 25.3 & 14.0 &14.6 &23.6 & 15.0 \\
         & \emph{Ours-Swin} & 23.4 & 33.7 & 17.6 & 24.5 & 25.4&15.2 &15.6 &26.5 &15.3 \\
         & \emph{Ours-Swin$^+$} & 22.8 & \textbf{34.6} & \textbf{18.5} & 24.8 & 26.4 & \textbf{15.5} & 15.0 & \textbf{27.7} & \textbf{16.2}\\
         
         \arrayrulecolor{black}\cmidrule{1-11}
         
         %--------------Test set---------
         \multirow{8}{*}{\rotatebox{90}{Test}}  & chance & 6.2 & 2.3 & 0.1 & 8.1 & 3.3 & 0.3 & 1.9 & 0.7 & 0.0 \\
         & \color{gray}AVT+ \cite{girdharAnticipativeVideoTransformer2021} & \color{gray}25.6 & \color{gray}28.8 & \color{gray}12.6 & \color{gray}20.9 & \color{gray}22.3 & \color{gray}8.8 & \color{gray}19.0 & \color{gray}22.0 & \color{gray}10.1 \\
         
         \arrayrulecolor{lightgray}\hhline{*{1}{~}*{10}{-}}\arrayrulecolor{black} \rule{0pt}{2.6ex}
         & RULSTM \cite{furnariWhatWouldYou2019} & \textbf{25.3} & 26.7 & 11.2 & 19.4 & 26.9 & 9.7 & \textbf{17.6} & 16.0 & 7.9 \\
         & TempAgg \cite{senerTemporalAggregateRepresentations2020} & 21.8 & 30.6 & 12.6 & 17.9 & 27.0 & 10.5 & 13.6 & 20.6 & 8.9 \\
         & TCN-TSN \cite{zatsarynnaMultiModalTemporalConvolutional2021} & 20.4 & 26.6 & 10.9 & 17.9 & 26.9 & 11.1 & 11.7 & 15.2 & 7.0 \\
         & TCN-TBN \cite{zatsarynnaMultiModalTemporalConvolutional2021} & 21.5 & 26.8 & 11.0 & \textbf{20.8} & \textbf{28.3} & \textbf{12.2} & 13.2 & 15.4 & 7.2 \\
         & \emph{Ours-TSN$^+$} & 19.4& 28.3&13.4 &14.0 &24.2 &9.9 &12.0 &19.5 &10.9 \\
         & \emph{Ours-Swin$^+$} & 20.7 &\textbf{31.8} &\textbf{14.9} &16.2 &27.7 &12.1 &13.4 &\textbf{23.8} &\textbf{11.8} \\
         \bottomrule
    \end{tabular}}
    \caption{Comparison of state-of-the-art methods on the validation and test set of EpicKitchens-100. Our models set a new state of the art. The numbers in bold-face indicate the highest score. All methods use all modalities provided by~\cite{furnariWhatWouldYou2019}, except for MeMViT which uses RGB only. TempAgg and the ones marked with $^+$ additionally use interacting hand-object bounding boxes and audio, respectively.}
    \label{tab:ek100_sota}
\end{table}

%% file: tables/results_egtea.tex
\begin{table}[t]
    \centering
    \resizebox{\linewidth}{!}{
%\small
\setlength\tabcolsep{3pt}
    \begin{tabular}{lcc>{\columncolor{Gray}}ccc>{\columncolor{Gray}}c>{\columncolor{Gray}}c}
        \toprule
        %Input
          \multirow{2}{*}{Method} & \multicolumn{3}{c}{Top-1} &  \multicolumn{3}{c}{Class mean @1} & \multicolumn{1}{c}{Top-5} \\
         \arrayrulecolor{gray} \cmidrule(lr){2-4} \cmidrule(lr){5-7} \cmidrule(lr){8-8} \arrayrulecolor{black}
         
         & Verb & Noun & Act. & Verb & Noun & Act. & Act. \\
         \midrule
         \color{gray}I3D-Res50 \cite{carreiraQuoVadisAction2017} & \color{gray}48.0 & \color{gray}42.1 & \color{gray}34.8 & \color{gray}31.3 & \color{gray}30.0 & \color{gray}23.2 & \color{gray}-\\
         \color{gray}FHOI \cite{liuForecastingHumanObjectInteraction2020} & \color{gray}49.0 & \color{gray}45.5 & \color{gray}36.6 & \color{gray}32.5 & \color{gray}32.7 & \color{gray}25.3 & \color{gray}- \\
         \color{gray}AVT \cite{girdharAnticipativeVideoTransformer2021} & \color{gray}54.9 & \color{gray}52.2 & \color{gray}43.0 & \color{gray}49.9 & \color{gray}48.3 & \color{gray}35.2 &\color{gray}-\\
         
         \arrayrulecolor{gray}\hhline{*{8}{-}}\arrayrulecolor{black} \rule{0pt}{2.6ex}

         RULSTM \cite{furnariWhatWouldYou2019} & - & - & - & - & - & - & 71.84$^\star$ \\
         ImagineRNN \cite{wuLearningAnticipateEgocentric2021} &- &- &- &- &- &- & 72.32$^\star$ \\
         AVT (TSN) \cite{girdharAnticipativeVideoTransformer2021} & 51.7 & 50.3 & 39.8 & 41.2 & 41.4 & 28.3 & -\\
         \sname-TSN (Ours) & \textbf{53.4}& \textbf{50.4}& \textbf{42.5}& \textbf{42.4}& \textbf{44.5}& \textbf{35.2}& \textbf{72.47}$^\star$\\ 
         \bottomrule
    \end{tabular}}
    \caption{Comparison to the state-of-the-art methods on EGTEA Gaze+ with $\tau_a=0.5s$. Results marked with $^\star$ are averaged across the three official splits, while others are based on split 1 only. We use the same input modalities as RULSTM. More details on the used modalities of each method can be found in the supplementary material.}
    \label{tab:egtea_sota}
\end{table}

%% file: sections/conclusion.tex
%auto-ignore
\section{Conclusion and Future Work}
This work presents \lsname, an attention based multi-modal feature fusion method for action anticipation. Extensive ablations demonstrate the improved performance of our approach compared to basic score fusion or other multi-modal fusion methods and in state-of-the-art comparisons AFFT outperforms existing approaches on EpicKitchens-100 and EGTEA Gaze+. Our method can easily be combined with various feature extractors and is extensible to new modalities without architectural changes. Given this extensibility we hope to provide a framework for multi-modal action anticipation for other researchers and aim to experiment on additional modalities like body poses and object hand interactions ourselves, in the future.

%% file: sections/supplementary.tex
%auto-ignore
\section*{Supplementary Material}
\renewcommand\thesubsection{\Alph{subsection}}
\renewcommand{\theequation}{S\arabic{equation}}
\renewcommand{\thefigure}{S\arabic{figure}}
\renewcommand{\thetable}{S\arabic{table}}
In this supplementary, we provide additional experiments and evaluations which did not fit in the main paper.

\subsection{Temporal and modality contributions}
\begin{figure}[b]
    \centering
        \includegraphics[width=\linewidth]{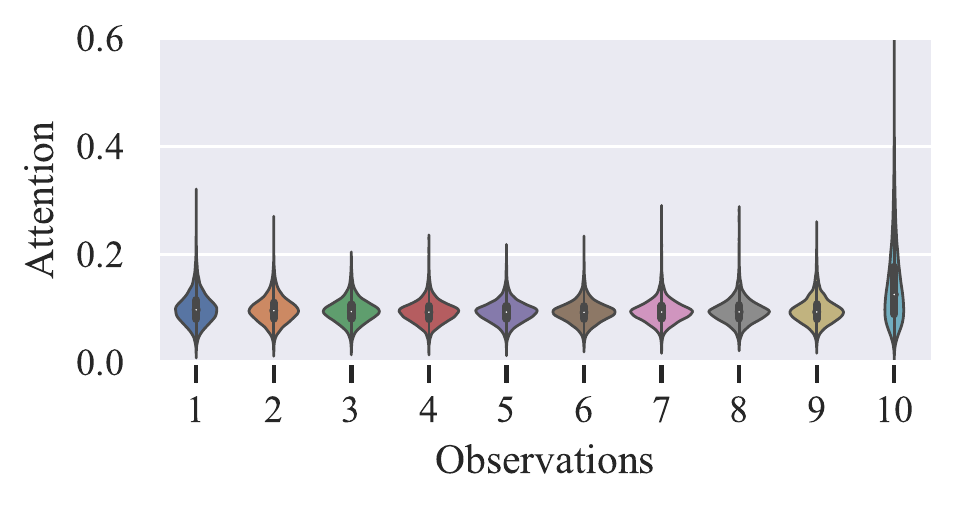}
        \caption{Temporal attentions of AFFT-TSN over all samples of the validation set of EpicKitchens-100. This figure describes a similar pattern and validates the evaluation on Swin features in figure~\ref{fig:temporal_attention_swin}.}
        \label{fig:temporal_attention_tsn}
\end{figure}

In Section~\ref{sec:temporal-context}, Figure~\ref{fig:temporal_attention_swin} displays temporal attention values for AFFT-Swin. In Figure~\ref{fig:temporal_attention_tsn} we show the same evaluation for the extracted TSN features. 

\begin{figure}[b]
    \centering
    \includegraphics[width=.7\linewidth]{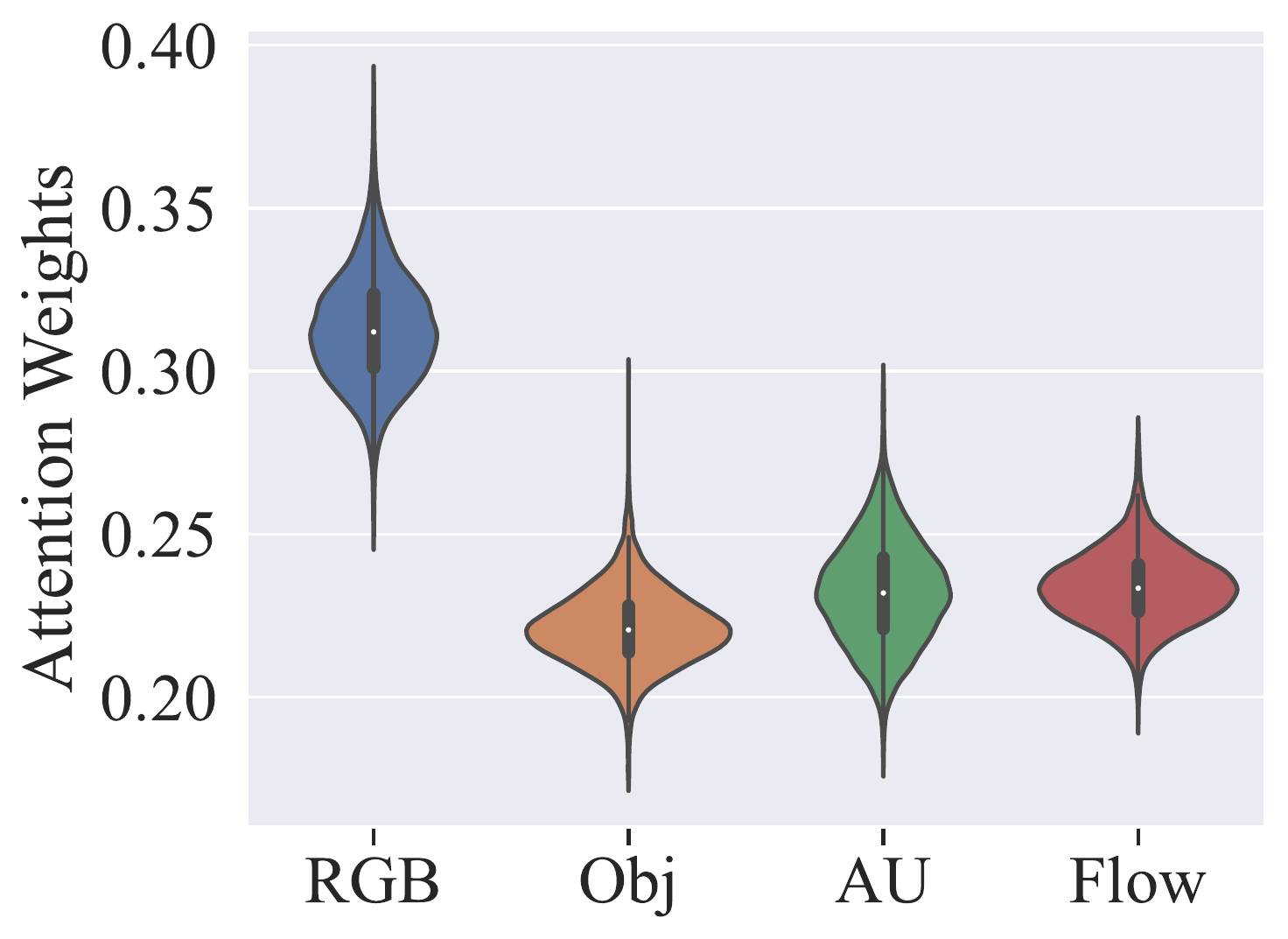}
    \caption{Modality attentions of AFFT-TSN on the validation set of EpicKitchens-100. This figure is the counterpart to Figure~\ref{fig:modality_attention_swin}, which describes the same evaluation on Swin RGB features.}
    \label{fig:modality_attention_tsn}
\end{figure}

Likewise, Figure~\ref{fig:modality_attention_swin} displays distributions of attention values over modalities using RGB-Swin features which is completed by Figure~\ref{fig:modality_attention_tsn} which displays the the same results for RGB-TSN features. 

\begin{figure*}[t]
    \centering
    \includegraphics[angle=90,height=1.2\linewidth]{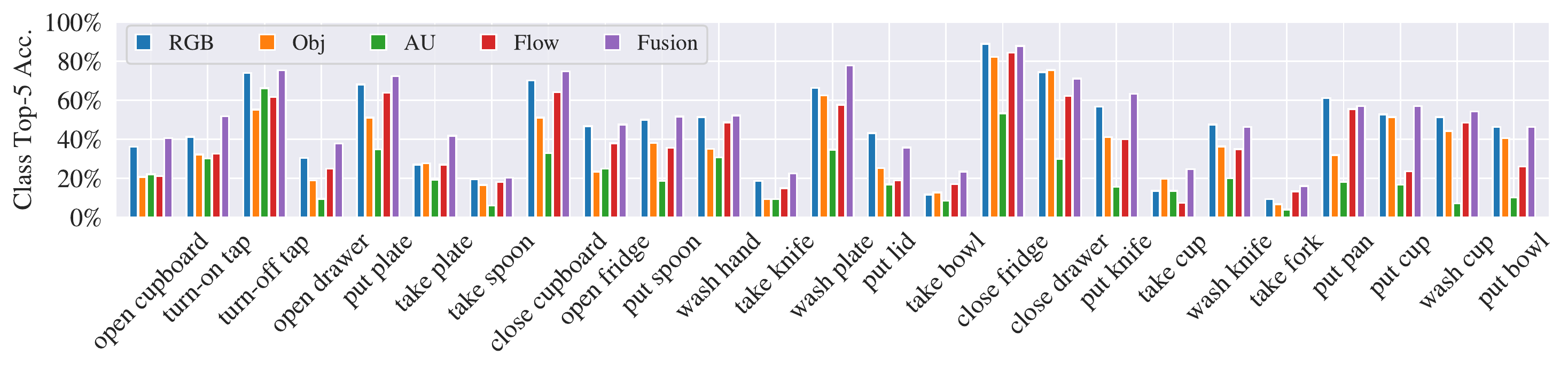}
    \caption{Per-class top-5 accuracy of fusion (\sname-Swin) and single modalities for the largest-25 actions in the validation set of EpicKitchens-100. The classes are presented in the order of sample frequency, from left to right. For most classes, the fusion method provides superior results to the single modalities.}
    \label{fig:class_acc_swin}
\end{figure*}

Figure~\ref{fig:class_acc_swin} shows per-class top-5 accuracy for the 30 action classes with most samples in the EpicKitchens-100 validation set, based on RGB-Swin Features. A similar chart is displayed in Figure~\ref{fig:class_acc_tsn} for TSN features. Note, that for such high frequent classes, performance is significantly higher than in the overall dataset. Still, our method does not only perform well for high frequent classes, but also shows significantly improved results for tail classes, as can be seen in Table~\ref{tab:ek100_sota}.

\subsection{Confusion Matrix}
\label{suppsec:confusion}
\begin{figure}[t]
    \centering
    \begin{subfigure}{0.49\linewidth}
        \includegraphics[width=\linewidth]{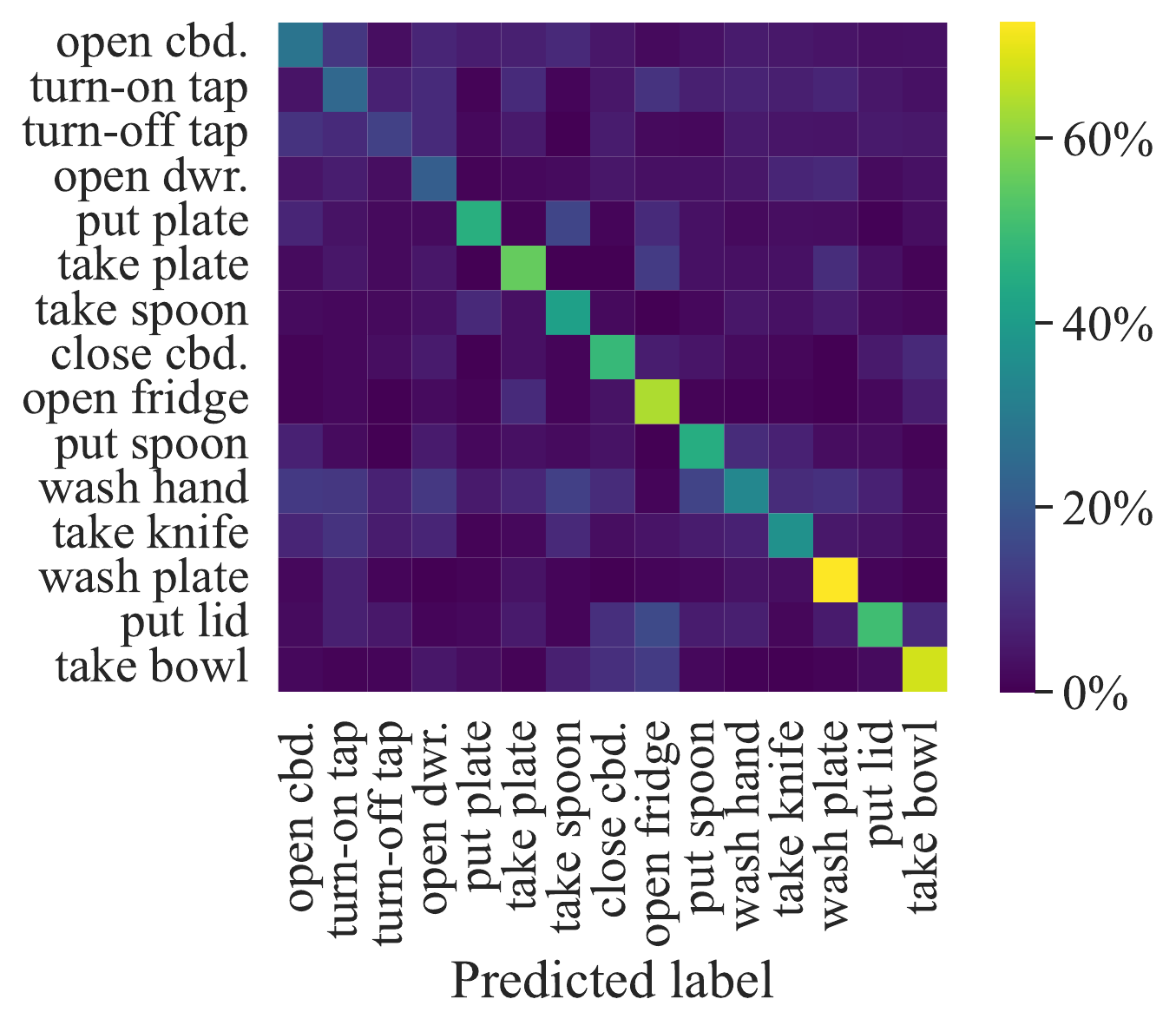}
    \end{subfigure}
    \begin{subfigure}{0.49\linewidth}
        \includegraphics[width=\linewidth]{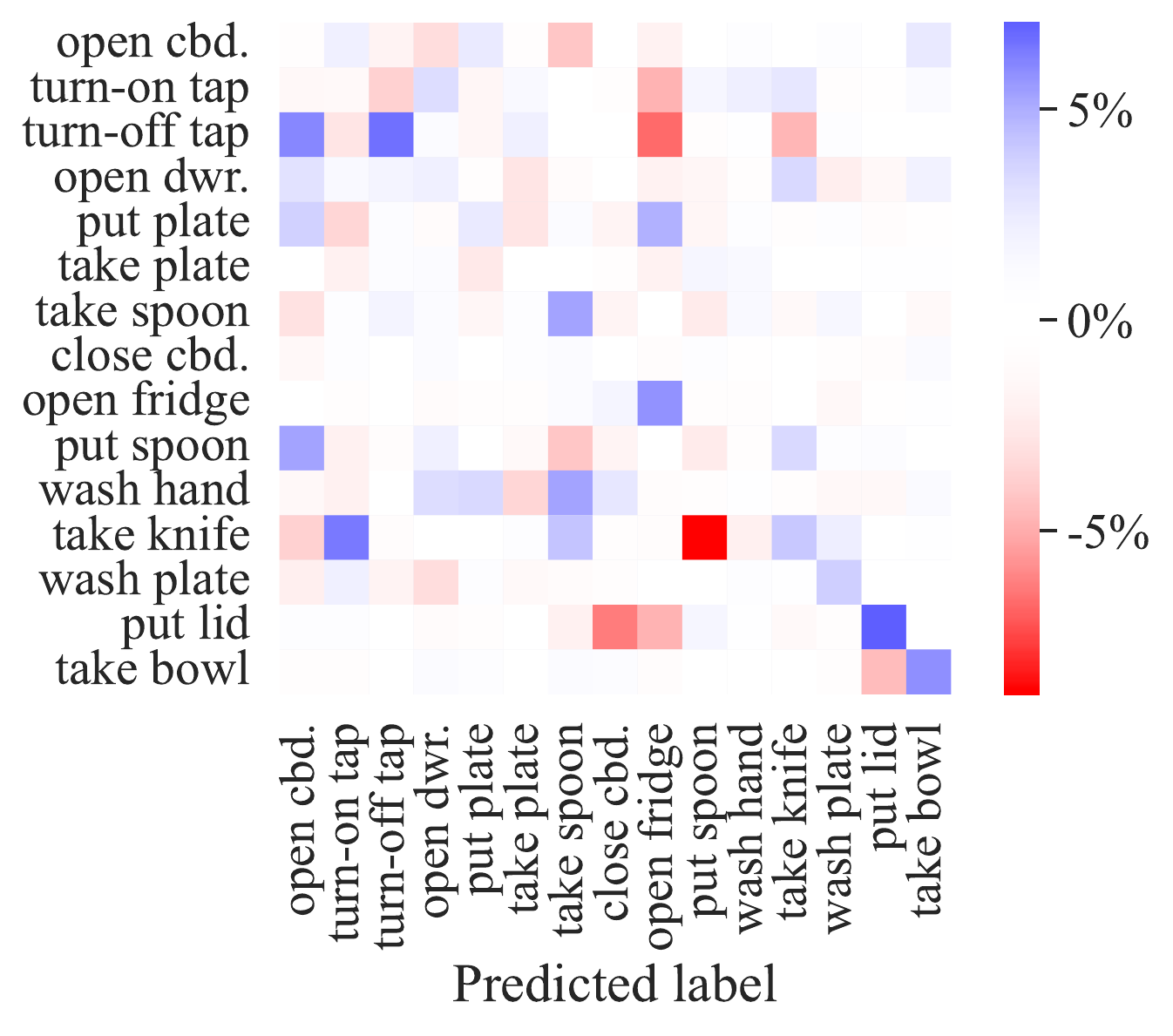}
    \end{subfigure}
    \begin{subfigure}{0.49\linewidth}
        \includegraphics[width=\linewidth]{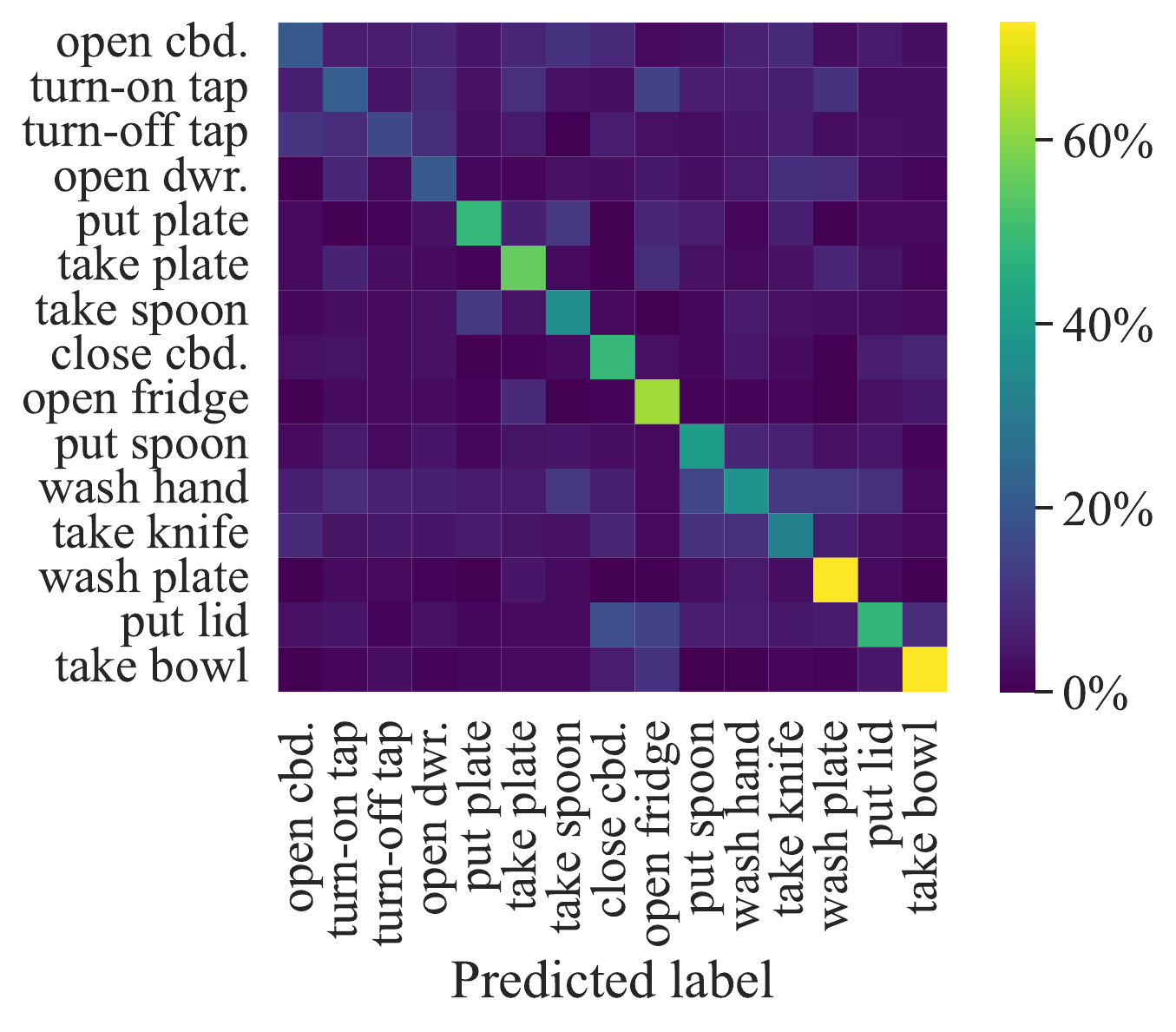}
    \end{subfigure}
    \begin{subfigure}{0.49\linewidth}
        \includegraphics[width=\linewidth]{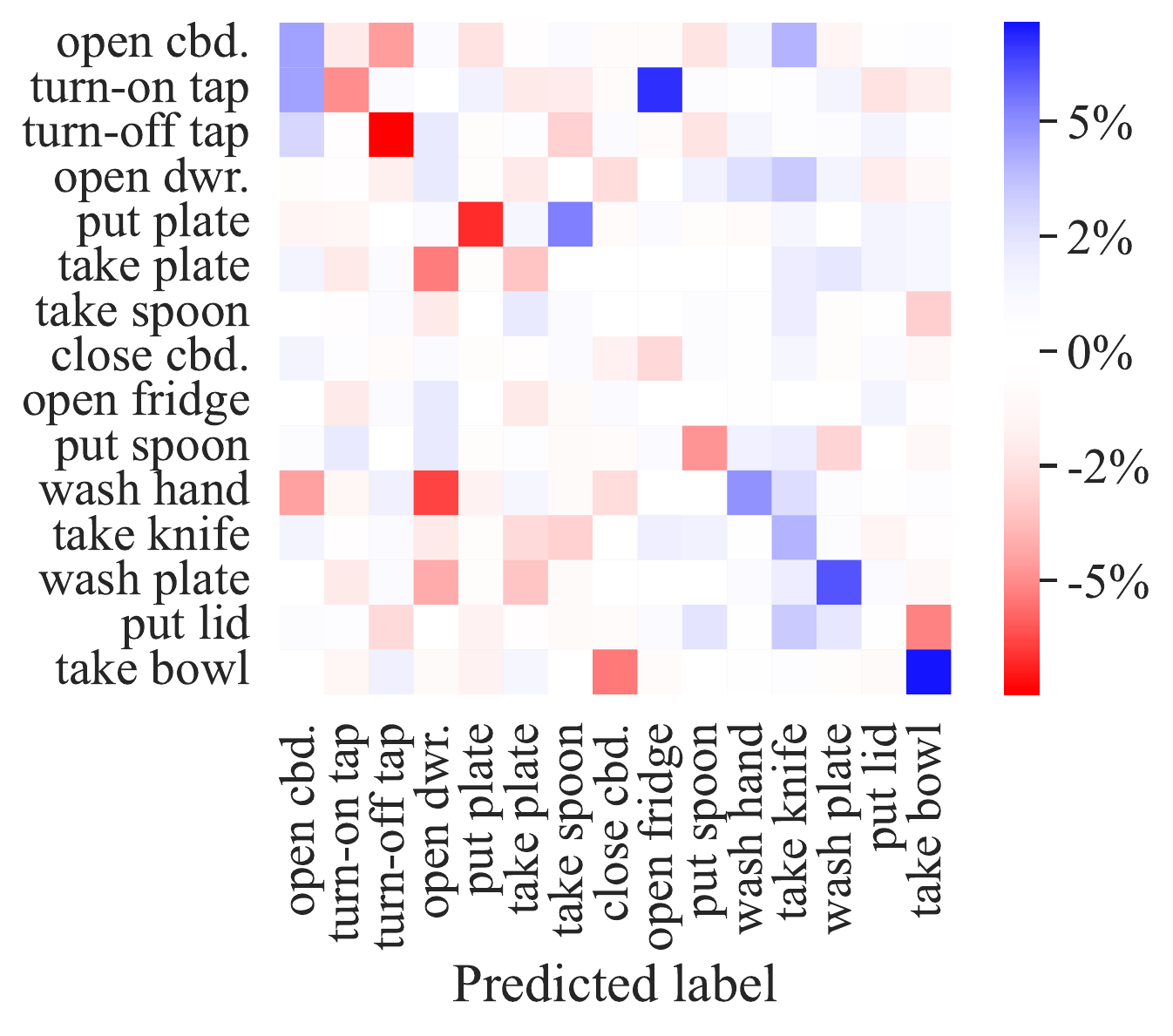}
    \end{subfigure}
    \caption{Confusion matrix for the largest-15 action classes in the validation set of EpicKitchens-100, with audio (left), as well as the difference to the confusion matrix without audio (right). From top to bottom, results of \sname-TSN and \sname-Swin are shown. An increase (blue) in confidence along the diagonal, especially obvious in the upper right figure, demonstrates the benefit of audio modality for egocentric action anticipation.}
    \label{fig:confusion}
\end{figure}

We follow the work of Kazakos et al.~\cite{kazakosEPICFusionAudioVisualTemporal2019} and evaluate the contribution of the audio modality, specifically. Figure~\ref{fig:confusion} shows the confusion matrices for the 15 most frequent action classes on the left and displays the difference to the confusion matrix without the audio modality on the right. While this Figure reflects the limited contributions of audio which are also visible in Table~\ref{tab:sub_fusion}, especially for Swin-Features, an increase of performance on the diagonal can be noted.

\subsection{Qualitative Results}
\label{subsec:qualitative}
\begin{figure}[t]
   \centering
     \begin{subfigure}[b]{\linewidth}
         \centering
         \includegraphics[width=\linewidth]{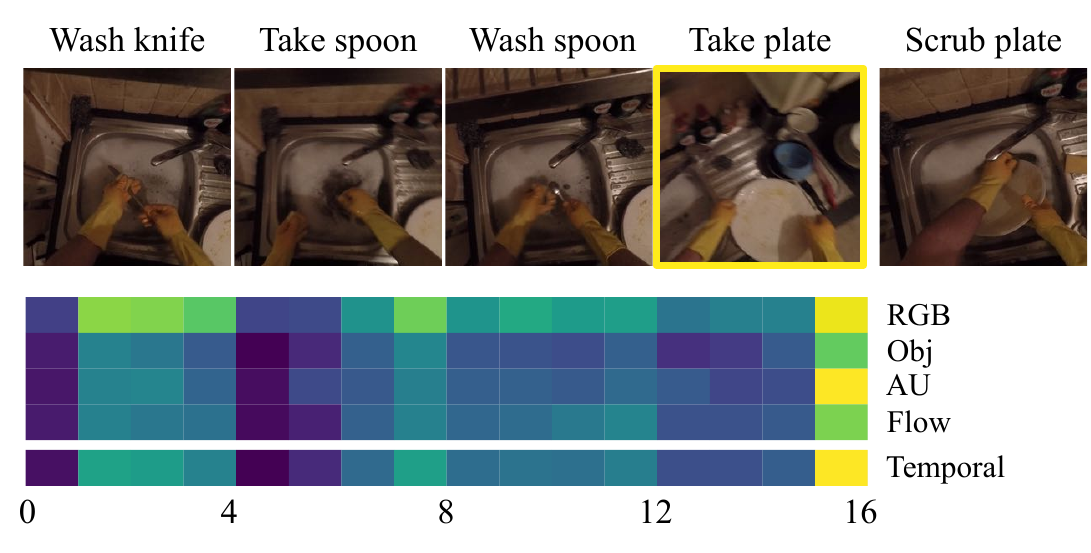}
         \caption{Future action: scrub plate.}
         \label{fig:sub_qualitative_plate}
     \end{subfigure}
     \begin{subfigure}[b]{\linewidth}
         \centering
         \includegraphics[width=\linewidth]{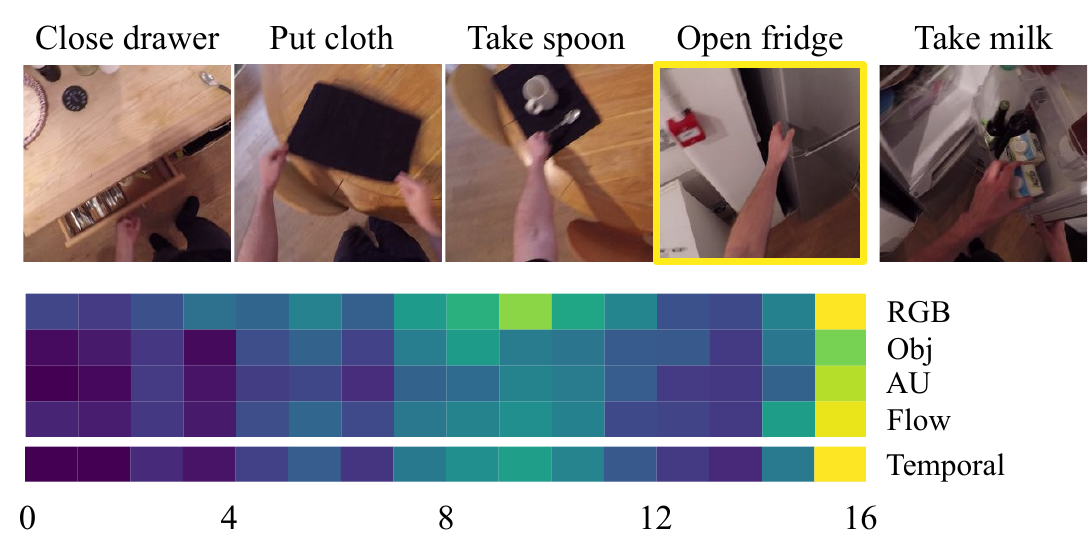}
         \caption{Future action: take milk.}
     \end{subfigure}
     \begin{subfigure}[b]{\linewidth}
         \centering
         \includegraphics[width=\linewidth]{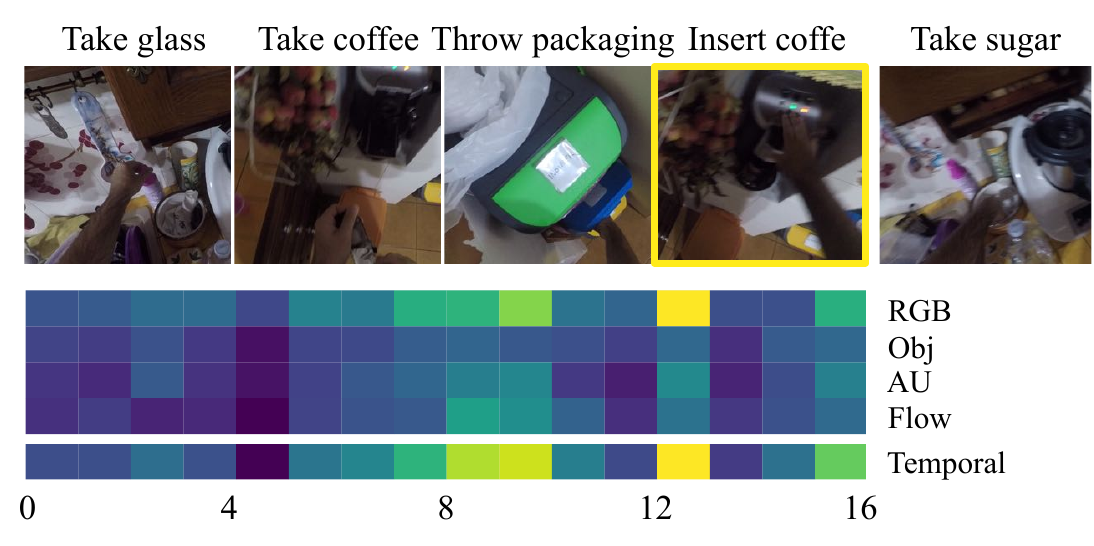}
         \caption{Future action: take sugar.}
     \end{subfigure}
     \begin{subfigure}[b]{\linewidth}
         \centering
         \includegraphics[width=\linewidth]{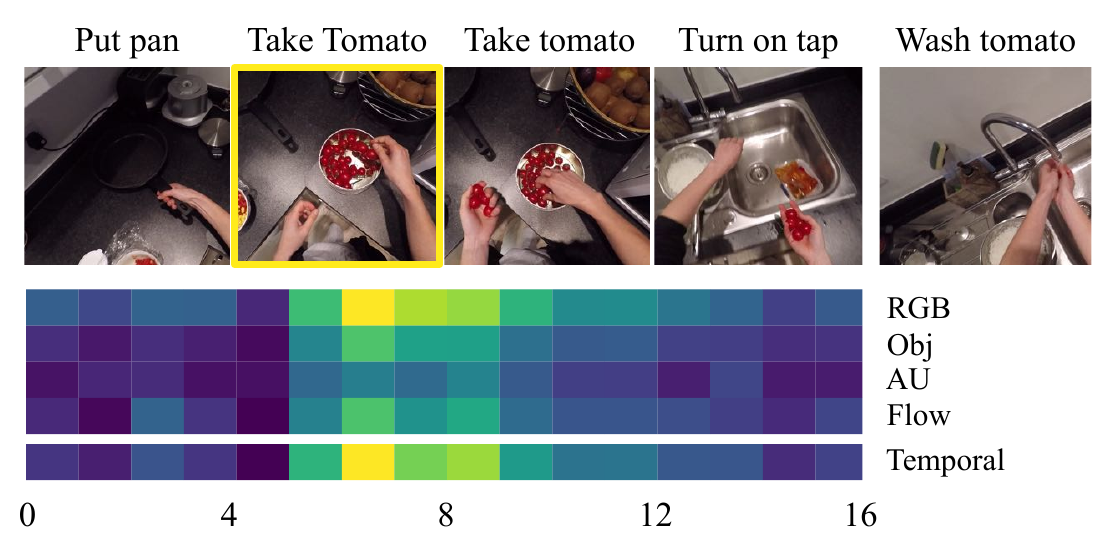}
         \caption{Future action: wash tomato.}
     \end{subfigure}
     \caption{Qualitative results on EpicKitchens-100. The horizontal and vertical axes indicate the index of the past frames and the modality as well as temporal attention scores, respectively. The closer the color is to yellow, the higher the attention score. We highlight a video frame with a yellow box when the attention score of the frame is highly activated.}
     \label{fig:qualitative}
\end{figure}

We plot additional visualization results of modality and temporal attentions in Figure~\ref{fig:qualitative}. The model used to generate such plots corresponds to the \sname-Swin in Table~\ref{tab:ek100_sota}. Each subfigure contains sampled frames showing temporal action evolution and modality and temporal attention map visualizations below. The frame receiving the most temporal attention is highlighted with a yellow box. From this experiment, we find that the proposed method attends dynamically to the multi-modalities and different past time steps to predict the future action, which demonstrates that our method successfully leverages long-term dependencies using multi-modal information for key frame detection and action anticipation.

\input{tables/supplementary_challenge}

\subsection{EpicKitchens-100 challenge}

Table~\ref{tab:ek100_challenge} lists results from the EpicKitchens-100 action anticipation challenge. This table relates to the test results in Table~\ref{tab:ek100_sota}. Entries to the challenge typically significantly surpass single-method performances, since it is common to ensemble differently trained models or results from different methods. We list this table separately, since its ensembled results can not be directly compared and did not undergo peer review.

\subsection{Details of used modalities on EGTEA Gaze+}
While a single RGB modality is used for I3D-Res50, FHOI~\cite{liuForecastingHumanObjectInteraction2020} adopts intentional hand movement as a feature representation, and jointly models and predicts the egocentric hand motion, interaction hotspots and future action. On the other hand, RULSTM~\cite{furnariWhatWouldYou2019} and ImagineRNN~\cite{wuLearningAnticipateEgocentric2021} make use of multiple modalities, i.e., RGB and optical flow, to further improve the anticipation performance of the next action. We note that the modalities used in AVT are ambiguous. We follow RULSTM and ImagineRNN and use RGB and optical flow as the input modalities for our method.

%% file: tables/supplementary_challenge.tex
\begin{table}[hbtp!]
    \centering
    \setlength\tabcolsep{1.5pt}
        \resizebox{\linewidth}{!}{
    \begin{tabular}{lcc>{\columncolor{Gray}}c>{\enspace}ccc>{\enspace}ccc}
        \toprule
         \multirow{2}{*}{Method} & \multicolumn{3}{c}{Overall} & \multicolumn{3}{c}{Unseen Kitchen} & \multicolumn{3}{c}{Tail Classes}\\ 
         \arrayrulecolor{gray} \cmidrule(lr){2-4} \cmidrule(lr){5-7} \cmidrule(lr){8-10} \arrayrulecolor{black}
          & Verb & Noun & Act. & Verb & Noun & Act. & Verb & Noun & Act. \\ \midrule
         
         AVT++ \cite{girdharAnticipativeVideoTransformer2021} & 26.7 & 32.3 & 16.7 & 21.0 & 27.6 & 12.9 & 19.3 & 24.0 & 13.8 \\
         allenxuuu & 29.9 & 30.4 & 17.4 & 25.1 & 26.1 & 14.1 & 24.6 & 23.7 & 14.3 \\
         PCO-PSNRD & 30.9 & 41.3 & 18.7 & 25.7 & 35.4 & 16.3 & 25.0 & 35.4 & 16.1 \\
         ICL-SJTU & \textbf{42.0} & 35.7 & 19.5 & \textbf{33.4} & 26.8 & 15.9 & \textbf{41.0} & 33.2 & 16.9 \\
         NVIDIA-UNIBZ & 29.7 & 38.5 & 19.6 & 23.5 & 35.2 & 16.4 & 23.5 & 31.1 & 16.6 \\
         SCUT & 37.9 & \textbf{41.7} & \textbf{20.4} & 27.9 & \textbf{37.1} & \textbf{18.3} & 32.4 & \textbf{36.1} & \textbf{17.1} \\
         \bottomrule
    \end{tabular}}
    \caption{Current leaders in the EpicKitchens-100 action anticipation challenge. The numbers in bold-face indicate the highest score.}
    \label{tab:ek100_challenge}
\end{table}